\documentclass{article}
\usepackage{spconf,amsmath,graphicx}

\title{Domain-aware Neural Language Models for Speech Recognition}
\name{Linda Liu, Yile Gu, Aditya Gourav, Ankur Gandhe, Shashank Kalmane, 
Denis Filimonov, Ariya Rastrow, Ivan Bulyko}
\name{%
\begin{tabular}{@{}c@{}}
Linda Liu \qquad
Yile Gu \qquad
Aditya Gourav \qquad
Ankur Gandhe \\
Shashank Kalmane \qquad
Denis Filimonov \qquad
Ariya Rastrow \qquad
Ivan Bulyko 
\end{tabular}}
\address{Amazon Alexa}
\begin{document}
%\ninept
%
\maketitle
\begin{abstract}
    As voice assistants become more ubiquitous, they are increasingly expected to support and perform well on a wide variety 
    of use-cases across different domains.
    We present a domain-aware rescoring framework suitable for achieving domain-adaptation during second-pass rescoring in production settings. 
    In our framework, we fine-tune a domain-general neural language model on several domains, and use an LSTM-based domain classification
    model to select the appropriate domain-adapted model to use for second-pass rescoring. 
    This domain-aware rescoring improves the word error rate by up to 2.4\% and slot word error rate by up to 4.1\% on three individual domains -- shopping,
    navigation, and music -- compared to domain general rescoring. These improvements are obtained while maintaining
    accuracy for the general use case.  

\end{abstract}
\section{Introduction}
\label{sec:intro}

Voice assistants have become increasingly popular and are used
for an ever-expanding set of use-cases. For example, users can currently ask voice assistants to play music
(music domain),
obtain a business address (navigation domain), and buy an item (shopping domain). 
Often, these  automatic speech recognition systems consist of a separate language
model (LM) component in the first-pass used in conjunction with an acoustic model, and an optional LM in the second-pass for rescoring. 
These LM models are trained to estimate the probability of a sequence of words $P(w_{1}..w{n})$. 
While n-gram LMs do this by estimating the probability of each word, given the previous $n-1$ words,
neural LMs (NLMs) learn a distributed representation for words as well as the probability function for the word sequence in
context \cite{bengio2003neural}. This allows them to generalize estimates to unseen word sequences,
and for longer word histories. 
Both types of LMs can be trained on a variety of textual data sources, 
and are typically optimized for the general use case.

However, different usage \emph{domains} may differ significantly in their language statistics and a general LM
may not perform as well on new domains, or domains otherwise not well represented by the general use case
\cite{blitzer2007domain}. 
Training a single LM that performs well on many domains is challenging. Some recent approaches include using
attention to acoustic embedddings during decoding \cite{gandhe2020audio} or a compositional neural language model
that learns how to combine multiple component language models \cite{filimonov2020neural}.

In this work, we describe a domain-aware rescoring framework that can be used to address this in a production
setting in the second-pass. We fine-tune a domain-general LM on data from three domains (music, navigation, shopping), and show that these
models capture improvements in word error rate (WER) and slot WER (SlotWER) -- which measures the performance on certain
critical content words --  beyond the general LM.
We find this holds even when using neural models capable of capturing longer word histories
and more complex relationships than traditional n-gram LMs. 
We demonstrate an effective way to combine a classification model to determine which domain-adapted model to use in
second-pass rescoring. 
With this framework, we are able to
obtain 0.7\%-4.1\% improvement on domain WER and SlotWER.
The experimental results that we report use an experimental ASR system that does not reflect the performance of the current Alexa ASR system in production. 

\section{Previous work}

Previous work on LM adaptation has shown that incorporating contextual or domain-adapted knowledge in the LM can improve the performance of ASR
systems. 
One way to achieve this is by dynamically adjusting the weights of an
interpolated n-gram LM, based on the preceding text \cite{kneser1993dynamic}. 
The interpolated LM can consist of different LMs that are optimized for different dialog states or applications
\cite{wessel1999comparison}, topic/domain
\cite{kneser1993dynamic,raju2018contextual},  or decomposition of topic factors \cite{bellegarda2000exploiting,gildea1999topic}. 
These LMs can be trained separately, and then the appropriate mixture
selected at runtime. Additionally, a cache component, which maintains a representation of recently occuring words, can be combined with an interpolated n-gram model to adapt the model to a target domain, based on the recent history \cite{kuhn1990cache}.

More recently, neural adapation approaches have been used to adapt a LM to a target domain based on
non-linguistic contextual signals, such as the application at the time of the request \cite{ma2018modeling}, or learned topic vectors \cite{chen2015recurrent, mikolov2012context}. For example, \cite{chen2015recurrent} used topic representations obtained from latent dirichlet allocation to adapt an NLM for genres and shows in a multi-genre broadcast transcription task. Domain-adaptation can also be achieved via shallow-fusion, in which an external (contextually constrained) LM is
integrated during beam search \cite{bahdanau2016end}. 

Given the limited search space in the second-pass, we can use larger and more
complex models in the second-pass of ASR systems, resulting in additional WER improvements \cite{bengio2003neural}. In this approach, some set of
hypotheses (represented as an n-best list, lattice, or confusion network) generated by a domain-general first-pass model is then rescored by a domain-adapted
model. The LM scores obtained from the first pass model are updated using some combination of the first and second pass
LM scores. 

Recently, \cite{vu2019online} demonstrated the feasibility of selecting adapted models to use in a production setting, 
using a pipeline with a topic classifier to select the appropriate domain-adapted LM. They show that this
setup results in improvements beyond using the domain-general model alone. However, their second-pass LM is
a 3-gram model. Recurrent neural LMs consistantly outperform traditional n-gram
models, particularly on less frequent tail words, and are capable of capturing more complex relationship
between words \cite{bengio2003neural, jozefowicz2016exploring}, so it is unclear whether domain-adapted improvements
might still hold when using a stronger LM. 

In the next section, we present our domain-aware rescoring framework, in which we select a domain-aware
NLM for second-pass based on the first pass ASR output.

\section{Domain-aware rescoring framework}

\begin{figure}[htb]
\centerline{\includegraphics[width=9cm]{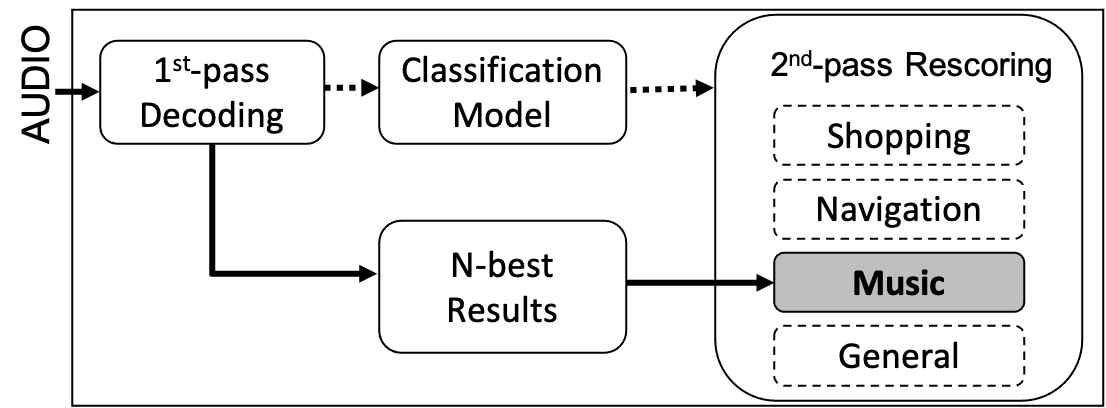}}
\caption{Second-pass rescoring framework. The classification model select
    the appropriate rescoring model based on the first-pass decoding output. The n-best hypotheses are then rescored by
    the selected model (e.g., music-specific model).}
\label{fig:diagram}
\end{figure}

Our proposed domain-aware rescoring framework is illustrated in Fig. \ref{fig:diagram}. After first-pass decoding, the
one-best ASR output is fed into a classification model, in order to determine which second-pass rescoring model to use. The
second-pass rescoring model is selected if the posteriors for a given class meet a selected threshold. The first and
second-pass LM scores are interpolated based on an optimization for minimizing WER. In the following sections, we
describe each of these components in turn: we first describe the first pass ASR system that we will use. Then we
describe the classification model used to select the second-pass rescoring model, based on the first pass output.
Finally, we describe the domain-adapted NLMs used in second-pass rescoring.

\subsection{First-pass model}
The first-pass LM is Kneser-Ney \cite{kneser1995improved} smoothed n-gram LM trained on a variety of data, including
in-domain data (transcribed and semi-supervised data from real user-agent interactions \footnote{The user data is anonymized and
consists of only a subset of data used for our production system}), and out-of-domain data (e.g., corpora available on the
web). It is optimized to minimize perplexity over general traffic, which covers all domains \cite{kneser1993dynamic}.
The acoustic model is a low-frame-rate model with 2-layer frquency LSTM \cite{li2017acoustic}
followed by a 5-layer time LSTM trained with connectionist temporal classification loss \cite{graves2006connectionist}.

\subsection{Classification model}
Selecting the appropriate second-pass rescoring model can be thought of as a text classification task. We train a
light-weight LSTM-based classifier. It consists of an input embedding layer with 100 units
and a vocabulary of roughly 150,000, including a token for out-of-vocabulary words, followed by an LSTM layer with 64 hidden units, a fully connected layer,
and an output layer with 4 output units. Each output unit corresponds to one of the second-pass models (one
domain-general model, three domain-adapted models). We use cross-entropy loss,
softmax activation, as well as the Adam optimizer with intial learning rate of 0.001 and early stopping based on
performance on a development set. 
At training time, the model is trained on annotated transcribed data. At test time, we feed in one-best ASR hypothesis.
Input utterances were padded or truncated to a length of 10. 

\subsection{Domain-adapted NLMs}
Following \cite{raju2019scalable}, we trained a NLM with two LSTM layers, each comprising 512 hidden
units. We use noise contrastive estimatation based training, which is
self-normalized and results in faster computation during training and inference, as it does not require the computation of the full softmax 
during training \cite{mnih2012fast}. To adapt the NLMs to individual domains, we initialized training from the converged domain-general model, using the subset of data from each of the
respective domains in order to fine-tune the LM to the domain \cite{biadsy2017}. Following
\cite{atkinson2018pseudo}, who used a smaller learning rate for fine-tuning, we used $1/4$ of the initial
leaning rate, which yielded the best improvements on perplexity (PPL) from a range of [$1/8$-$1/2$].
In this way, each of our NLMs was trained on the same data, and also used the same vocabulary.

\subsection{Second-pass rescoring and optimization}
We generate a lattice from the 10-best ASR hypotheses, and rescore it using the second-pass model selected by the domain
classifier. Rescoring itself uses the push-forward algorithm as described in \cite{auli2013joint, kumar2017lattice}. 
Additionally, we optimize the interpolation weights used to combine the first- and second-pass LMs based on (1) overall
WER and (2) SlotWER.
The optimization is carried out using the simulated annealing algorithm as
described in \cite{xiangGenSA}.

We compare these results to an offline method where we use the one best ASR results to estimate per-utterance interpolation
weights for each of our rescoring models using the EM algorithm. We then use these weights to do an linear interpolation
between the models in the second-pass \cite{kneser1993dynamic}. This approach is not practical in a production setting, but provides a
point of comparison for our results.

\section{Experiments}

\subsection{Second-pass models}
\begin{table}[h]
    \begin{tabular}{lrrrr}
        Model & PPL$_{Other}$ & PPL$_{Nav}$ & PPL$_{Music}$ & PPL$_{Shop}$\\
        \hline
        LM$_{Genrl}$ & 0\% & 0\% & 0\%  & 0\%  \\ 
        LM$_{Nav}$ & 917.1\% & \bf{-47.7\%} & 1521.9\% & 314.5\% \\
        LM$_{Music}$ & 331.7\% &734.3\%&  \bf{-43.6\%} & 562.5\%\\
        LM$_{Shop}$ & 493.4\% & 349.6\% & 742.6\% & \bf{-45.8\%} \\
    \end{tabular}
    \caption{Relative PPL on evaluation set by domain split each domain-adapted LM compared to the domain-general LM
    baseline}
    \label{table:ppl}
\end{table}
The domain-general NLM was trained on 80 million utterances from anonymized user interactions with Alexa, constituting a portion of
all live traffic. A portion of these utterances further included annotations for their domains. Among those utterances that were annotated, 
a portion were navigation, music, and shopping domain utterances (over 4 million each). We selected the top 238k most frequent words from this corpus as the vocabulary for the NLM, and mapped the out-of-vocabulary 
tokens to $<$unk$>$. The $<$unk$>$ token was scaled by a factor of $10^{-5}$ in the rescoring experiments. 

We adapted this domain-general NLM to three separate domains
by fine-tuning of the annotated data for that domain: music, navigation, and shopping. Table \ref{table:ppl} shows the perplexities for our test set, split out by domain, for each of the domain-adapted LMs and the domain-general LM. 
The \emph{Other} partition does not contain music, navigation, shopping domain utterances, and only includes other
domains, such as factual questions and commands to control connected home appliances.
We observe a reduction in PPL on the evaluation data for the corresponding domain
when using the domain-adapted model as compared to the domain-general model. 
However, we also observe an increase in the PPL of the \emph{Other} partition when using any domain-adapted model. 
This suggests that conditioning the second-pass LM on domain of the utterance can improve recognition, and supports the
use of a domain classifier.

\subsection{Domain classification model}

\begin{table}[h]
    \begin{tabular}{rrrrr}
        Domain & Music & Navigation & Shopping & Other \\
        \hline
        Precision & 0.94 & 0.92 & 0.92 & 0.97 \\
        Recall & 0.93 & 0.89 & 0.97 & 0.92 
    \end{tabular}
    \caption{Per domain precision and recall for model classifier used to select the second-pass rescoring model}
    \label{table:classifier_metrics}
\end{table}

This model is trained with roughly 9 million utterances from user interactions with Alexa that were human-annotated
for domain, using a 8-1-1 split for dev and eval (a subset of the total annotated data available to us).
All utterances are first-turn interactions, in which the
user directly prompts Alexa, and does not contain utterances that consist of only the wakeword (e.g., \emph{``Alexa''}).

Across all domains, we observe an overall classification accuracy
of 95.9\%, with unweighted precision of 93.9\% and recall of 92.8\%, based on the max class. 
Per-domain classification breakdown is shown in Table \ref{table:classifier_metrics}. In our rescoring framework, we
select the domain-adapted model for rescoring only in cases where the posteriors for that class meet some
specified threshold (the higher the threshold, the more conservative the classification).
Based on precision and recall values per domain, we used a 0.85 threshold across domains, resulting in an overall
classification accuracy of 91\%.

\begin{table*}[]
  \centering
    \begin{tabular}{lrrrrrrr}
        Model & \multicolumn{2}{c}{Navigation} & \multicolumn{2}{c}{Music} & \multicolumn{2}{c}{Shopping} & Other\\
              & WER & SlotWER & WER & SlotWER & WER & SlotWER & WER\\
        \hline
        LM$_{Genrl} $ & -1.3\% & -3.4\% & -0.6\% & 1.0\% & -2.2\% & -0.3\% & -0.8\% \\
        \hline
        LM$_{Nav}$ & -5.3\% & -7.3\% & -- & -- & -- & -- & 1.8\% \\
        LM$_{Music}$ & -- & -- & -2.6\% & -4.1\% & -- & -- & 2.5\% \\
        LM$_{Shop}$ & -- & -- & -- & -- & -4.0\% & -4.1\%  & 1.7\% \\
        \hline
        DomainAware & -3.7\% & -6.5\% & -2.3\% & -3.1\% & -2.9\% & -3.6\% & -0.9\% \\
        AdaptationBaseline & -1.6\% & -4.5\% & -3.2\% & -4.7\% & -1.7\% & -2.7\% & -0.9\%  \\
        \hline
       Oracle & -31.2\% & -- & -25.2\% & -- & -32.8\% & -- & -18.4\%
    \end{tabular}
    \caption{WER improvements (shown as negative number) on domain-adapted test sets using general (LM$_{Genrl}$), domain-adapted LM (LM$_{Nav,
    Music, Shop}$), domain-aware rescoring with a domain classifier (DomainAware), or the dynamically interpolated model
    (AdaptationBaseline) as a second-pass rescorer. All relative comparisons are respect to the first-pass baseline (not
    shown).
The Oracle provides a lower bound for the minimum WER achieveable, given a second pass model that always
prefers the hypothesis in the nbest list with the lowest WER.
}
    \label{table:wer}
\end{table*}

\subsection{Evaluation Data}
All of our experiments are evaluated on an evaluation set consisting of anonymized live user interactions with Alexa. The evaluation set consists
of annotated data across several different Alexa devices for a select time period. 
The final evaluation set consists of roughly 10-20k utterances for each of the shopping, music, and navigation domains. The
remainder of the set consists of 115k utterances from other domains.
Additionally, for each domain, we annotate the popular contentful slots, in order to evaluate
improvements in the SlotWER. These slots are the SongName, ArtistName, and AlbumName slots for
Music, the PlaceName and StreetName slots for Navigation, and the ItemName slot for Shopping.
Improving on recognition of the slot content
is especially important for voice interactions, where the WER is higher. Slots are often specific to
certain domains, making them most likely to benefit from domain-aware rescoring.

We created an development set using the same criteria as our evaluation set. This development set was used to determine the interpolation weights used
to combine the LM scores from the first- and second-pass models. 

\subsection{Results}
We present WER results obtained using our domain-aware escoring framework (DomainAware), in which we rescore each utterance based on the model
selected by the domain classifier (Table \ref{table:wer}). 
We show these results split by each domain based on the annotation in our test set (Navigation, Music, Shopping), as
well on the other domains (Other). The \emph{Other} partition does not contain utterances from the aforementioned domains.
Compared to the LM$_{Genrl}$ alone, we find that domain-aware rescoring results show improvements of 0.7\%-2.4\% on WER and 3.1\%-4.1\% on SlotWER, across all domains.
We also find a minor improvement (0.1\% WERR) on the Other partition of the test set; these improvements are not obtained at the expense of general WER. 

We also present the results that can be obtained on each evaluation set split, when rescoring that
split using only
the corresponding rescoring LM (LM$_{Nav,Music,Shop}$). This shows that the domain-adapted LMs can improve WER beyond the domain-general LM, even when
trained using the same data. Compared to using LM$_{Genrl}$, we observe improvements of 1.8\%-4.0\% on WER and
3.8\%-5.1\% on SlotWER of the corresponding evaluation set split.
Taken together, this shows that domain-aware rescoring allows us to capture the most of the improvements from using domain-adapted models, all while
avoiding degradation on other domains.

We conducted the above comparisons using a weighted combination of the first and second pass models, with interpolation
weights between first and second-pass models obtained by optimizing for WER or SlotWER on our development set. We observed no significant impact based on whether we used WER or
SlotWER; consequently, all of the aforementioned results are obtained from WER optimization. 

Finally, we compare the results obtained using domain-aware rescoring to the standard approach of dynamically adapting 
the interpolation weights over LM$_{Genrl}$, LM$_{Nav}$, LM$_{Music}$, and LM$_{Shop}$. 
This is the AdaptationBaseline shown in Table \ref{table:wer}. From this, we see that the
DomainAware approach is competitive with the dynamically interpolated adaptation baseline  -- both improve over the general neural language model,
but DomainAware shows larger WER improvements for the navigation and shopping domains, and smaller WER
improvements in the music domain. %TODO: check if interpolation weights have more weight to general% 
Though the DomainAware approach uses a domain classifier to select \emph{one} model for rescoring, the single model used is trained to
minimize perplexity on the respective domain; this is similar to the AdaptationBaseline, in which the interpolation weights are determined to minimize
the perplexity for the word sequence, using a combination of all the models. However, unlike the DomainAware approach,
the AdaptationBaseline approach is not practical for second-pass rescoring in production due to the latency
incurred during weight optimization. %TODO UPDATE ONCE WE SEE RESULTS

\section{Conclusion}
In this work, we present a framework for domain-aware rescoring that is suitable for use in a production
environment. We show that using this framework enables us to capture WER improvements on domains beyond what is captured by a
domain-general model trained on the same data. This framework has the benefit of allowing individual rescoring models to
be maintained independently,
making it suitable for asynchronous model updates. We demonstrate that domain-aware rescoring is competitive with traditional model
interpolation methods, which can only be run in an offline environment when latency is not an issue. One additional benefit of a separate classification model is that
 individual domain-adapted models can be maintained and updated separately. 
Future work looks to combine the classification model and second-pass rescoring models into a single model; i.e., by
incorporating an additional domain embeddding during the training of the NLM. The model would allow the NLM to
dynamically adapt a specific domain, and can be further extended to other characteristics of the user or utterance
beyond domain.

%\vfill\pagebreak
\clearpage

\bibliographystyle{IEEEbib}
\bibliography{refs}

\end{document}